\documentclass[10pt,letterpaper]{article}
\ifdefined\pdftexversion\pdfoutput=1\fi 
\usepackage[margin=1in]{geometry}
\usepackage{graphicx}
\usepackage{float}  
\usepackage{booktabs}
\usepackage{amsmath}
\usepackage{microtype}
\usepackage[T1]{fontenc}
\usepackage{lmodern}
\usepackage{tikz}
\usetikzlibrary{arrows.meta, positioning, fit, backgrounds}
\usepackage[colorlinks=true, allcolors=blue!60!black]{hyperref}
\usepackage{caption}
\usepackage{titlesec}
\titlespacing*{\section}{0pt}{1.5ex plus .3ex}{0.8ex}
\titlespacing*{\subsection}{0pt}{1.2ex plus .3ex}{0.5ex}
\newcommand{\code}[1]{\texttt{\small #1}}

\title{\vspace{-2em}\bfseries mbariml: a curation pipeline for turning deep-sea
imagery and video into object-detection training data}
\author{Lonny Lundsten \quad Kevin Barnard \quad Dave Caress\\[2pt]
\normalsize Monterey Bay Aquarium Research Institute\\
\normalsize \texttt{lonny@mbari.org}}
\date{\normalsize\today}

\begin{document}
\maketitle

\begin{center}\small
Source code, documentation, and issue tracker:\\
\url{https://github.com/mbari-org/mbari-ml}
\end{center}

\begin{abstract}
\noindent
Training data quantity and quality greatly affect object detection model performance, regardless of model architecture. For object detection in deep-sea video and imagery, where the objects of interest---primarily organisms---are sparse, faint, and hard to identify, incremental improvements to detector performance may require an iterative approach to data labeling and management. This paper presents \code{mbariml}, a Python-based video and image analysis pipeline built around the data labeling and management process. \code{mbariml} uses an Ultralytics YOLO detection model, runs it over still images or video, stores every detection as a reviewable region of interest, groups those regions by visual similarity so that a human can accept or reject them in bulk, and exports the result as training data, statistics, image sidecars, and additional metadata. Existing YOLO and Pascal VOC datasets can be imported into the same database, so a legacy training set can be reviewed, extended, and re-exported alongside new detections. The human review stage is the center of the design: an annotator can validate, relabel, resize, delete, and draw entirely new localizations, optionally assisted by the SAM3 segmentation model, and every one of those edits is written back to the same database the detector wrote to. Video receives particular attention: the software treats each tracker-produced track as a provisional observation and selects one representative frame instead of retaining every detection in the track. We describe the pipeline stage by stage, including how each track's representative frame is chosen---from a user-selected third of the track, the middle by default---which we examine on 684 tracks from seafloor video.
\end{abstract}

\section{The problem}

Deep-sea image and video surveys are collected from autonomous, remotely operated, landed, and crewed platforms. Regardless of the platform, an imaging system collects video or still images that need to be analyzed for the presence of organisms. The sheer volume of data that can be collected on each deployment makes manual annotation a bottleneck to analysis and discovery. Modern machine learning approaches can speed up this process by automatically detecting and localizing objects of interest, but they require high-quality labeled training data to perform well. While model architecture is advancing rapidly, useful workflows and tools that balance human validation effort and machine assistance are essential.

While object detection and tracking can greatly improve the speed at which these data can be analyzed, they introduce problems of their own. Run a model at a confidence threshold low enough to rarely miss potential objects of interest, and it returns a large number of false positives that must be manually edited or rejected. Run the model conservatively, and it silently misses the things that should not be missed. Either approach generates an enormous number of candidate detections that have to be validated by human experts.

Our workflow deliberately over-detects, running models with a low confidence threshold so that important detections are not missed. Image embeddings are calculated using DINOv3 for each region of interest (ROI) and stored as vectors in the database. The results can then be grouped in two ways. First, the grid of ROIs can be sorted by embedding similarity (cosine distance). Second, a clustering algorithm can optionally be applied to group ROIs into clusters of visually similar detections and auto-label them based on the cluster majority label. In this way, hundreds of similar detections can be edited or validated in batch. When the initial object detection model is a general, single-class detector (one whose only class is \code{object}, for example), the auto-labeling step names clusters numerically: \code{object\_1}, \code{object\_2}, \code{object\_3}, and so on.

The following sections describe the software and workflow in detail. We start with an overview of the pipeline, followed by an explanation of each stage, highlighting key design decisions and operational considerations.

\section{What the pipeline does}

\code{mbariml} is a command-line tool with a graphical review interface, organized into five stages: Import, Generate, Enrich, Review, and Export
(Figure~\ref{fig:pipeline}). In practice a run looks like this:

\begin{enumerate}\itemsep2pt
\item \textbf{Import or Generate:} import an existing labeled YOLO or Pascal VOC
      dataset, or point a supported Ultralytics YOLO detection model at a folder
      of images or at video. Either way, every box becomes one row carrying its
      own cropped image.
\item \textbf{Enrich} (optional): embed and cluster those crops, then relabel a
      whole cluster at once using its dominant label.
\item \textbf{Review:} by hand, validate, relabel, adjust boxes, delete mistakes, and draw in the detections the model missed.
\item \textbf{Export:} write training data, summary statistics, or a browsable gallery.
\end{enumerate}

Every stage reads and writes the same DuckDB file, in which a single flat
table holds one row per region of interest. There is no
per-stage intermediate format. Stages may be skipped when the fields they need
are already present: for example, pointing the video command at new footage and
going straight to review is as valid as running the optional embedding and
clustering steps first. Embedding still requires regions, and clustering
requires embeddings.

Each row stores its ROI inline as a JPEG blob. Embedding, clustering, and the
review GUI never alter the original imagery; the full frame is read-only and is shown only in the review interface, as context for the reviewer.

A single convenience command chains the scriptable part of that sequence---\code{ingest}, \code{embed}, \code{cluster}, \code{export}---over images or video, and can start or stop at
any named step, so re-exporting after a review session does not rerun inference. The interactive and ad hoc commands deliberately stay outside that
chain.

The Import and Generate commands allocate row identifiers from the database's
own counter rather than from zero, so a second run appends to an existing file instead of colliding with
it. Several videos, or a mix of stills and video from one deployment, can
therefore accumulate in a single database and be clustered and reviewed
together---which matters because visual similarity is more useful across a whole
deployment than within one file.

\begin{figure}[H]
\centering
\begin{tikzpicture}[
  box/.style={draw=black!55, rounded corners=1.5pt, align=center, font=\footnotesize,
              minimum height=6mm, minimum width=26mm, inner xsep=3pt, fill=white},
  phase/.style={draw=black!30, rounded corners=2.5pt, inner sep=5pt, fill=black!2},
  lbl/.style={font=\scriptsize\bfseries, black!65},
  ar/.style={-{Stealth[length=2.2mm]}, black!55, thick}, x=0.95cm]
\node[box] (iy) at (0,0) {\code{import yolo}};
\node[box] (iv0) at (0,-0.85) {\code{import voc}};
\begin{scope}[on background layer]\node[phase, fit=(iy)(iv0)] (p0) {};\end{scope}
\node[lbl, above=1pt of p0] {IMPORT};
\node[box] (ii) at (3.35,0) {\code{infer images}};
\node[box] (iv) at (3.35,-0.85) {\code{infer video}};
\begin{scope}[on background layer]\node[phase, fit=(ii)(iv)] (p1) {};\end{scope}
\node[lbl, above=1pt of p1] {GENERATE};
\node[box, minimum width=21mm] (em) at (6.55,0.42) {\code{embed}};
\node[box, minimum width=21mm] (cl) at (6.55,-0.43) {\code{cluster}};
\node[box, minimum width=21mm] (rf) at (6.55,-1.28) {\code{refine}};
\begin{scope}[on background layer]\node[phase, fit=(em)(cl)(rf)] (p2) {};\end{scope}
\node[lbl, above=1pt of p2] {ENRICH};
\node[box] (rv) at (9.75,0) {\code{review} (GUI)};
\node[box] (rm) at (9.75,-0.85) {\code{remap-labels}};
\begin{scope}[on background layer]\node[phase, fit=(rv)(rm)] (p3) {};\end{scope}
\node[lbl, above=1pt of p3] {REVIEW};
\node[box, minimum width=29mm, align=center] (ex) at (13.3,0.1)
      {\code{export}\\ voc\,/\,yolo\,/\,id\,/\,html\,/\,stats};
\node[box, minimum width=29mm] (st) at (13.3,-1.0) {\code{query}};
\begin{scope}[on background layer]\node[phase, fit=(ex)(st)] (p4) {};\end{scope}
\node[lbl, above=1pt of p4] {EXPORT};
\draw[ar] (p1) -- (p2); \draw[ar] (p2) -- (p3); \draw[ar] (p3) -- (p4);
\node[draw=black!50, fill=black!5, rounded corners=2.5pt, font=\footnotesize,
      minimum width=150mm, minimum height=7mm, align=center] (db) at (6.65,-2.85)
      {one DuckDB file \;---\; one row per region of interest, crop stored inline};
\foreach \p in {p0,p1,p2,p3,p4}
  \draw[{Stealth[length=2mm]}-{Stealth[length=2mm]}, black!45, dashed] (\p.south) -- (\p.south |- db.north);
\end{tikzpicture}
\caption{The five stages. All of them read and write the same \code{predictions} table, so any
stage's output can be fed to any other stage that needs what it contains.
Import and Generate are alternative entry points, and either can be followed
directly by review.}
\label{fig:pipeline}
\end{figure}

\subsection{One row per region of interest}

The database holds two tables. Almost everything is in one flat table,
\code{predictions}, with one row per region of interest. Each row records where
the region came from (\code{image\_path}, and for video the \code{video\_path},
\code{frame\_number}, \code{frame\_time\_s}, \code{track\_id}, and \code{track\_length} that lead back to the source footage), what the detector
said (\code{x\_min}\ldots\code{y\_max}, \code{class\_id}, \code{confidence},
\code{label}), what was derived from it (\code{roi}, the JPEG crop;
\code{sharpness}, a Laplacian-variance blur score; \code{embedding}, the DINOv3
vector), and what a human decided (\code{new\_label}, \code{verified}). The
second table, \code{run\_info}, holds a single row naming the model that
produced the detections, so an export can state its own provenance without the
operator having to retype it.

Columns introduced after the first release are added by \code{ALTER TABLE ... ADD COLUMN IF
NOT EXISTS} rather than by editing the table definition, so that a database
written by an earlier version opens and migrates rather than failing. The
practical consequence is that video support, the verified flag, and the track
provenance columns were all introduced without invalidating existing curated
databases.

\subsection{Two columns carry the curation state}

The distinction between \code{label} and \code{new\_label} is the one piece of
the schema that most affects what a training export contains, and it is easy to
get wrong. \code{label} is the detector's proposal. \code{new\_label} is written
only when a reviewer \emph{changes} the name. \code{verified} is set whenever a
reviewer signs off on a region, whether or not they renamed it---and applying a
label sets both, since naming something is itself an act of review.

The consequence is that a region the detector already got right, which a
reviewer merely confirmed, carries \code{verified = 1} and a \code{new\_label}
that is still null. A region's effective label is therefore resolved as
\code{COALESCE(new\_label, label)}: the curated name where a reviewer typed one,
the detector's own name where they confirmed it unchanged. Selection is on
\code{verified}, so an export contains exactly what a human has signed off on,
whether or not they renamed it. The rule is defined once and shared by all five
export paths, so a summary of a database and an export of it cannot describe
different populations. Regions labeled \code{noise} are excluded throughout.

\section{Import: existing labeled datasets}

Not every curation effort starts from fresh detections. Training sets built
earlier, by hand or with other tools, are usually what a new effort most needs
to review and extend. The import stage reads a YOLO dataset (images, label
files, and a names file) or a Pascal VOC dataset (images and XML) and writes
every box as an ordinary row, crop and blur score included, so that review,
embedding, clustering, and every export treat it exactly like detector output.

Imported boxes are marked as verified by default, because a training set is
human-labeled ground truth and the exports select only verified rows;
predictions from another model can instead be imported unverified, for review.
Label files are paired with images by relative path, then by filename where
that is unique, and a VOC file's recorded absolute path is never trusted, since
it is correct only on the machine that wrote it. Every YOLO label file is
checked against the names file before anything is written, so the wrong names
file fails immediately rather than mislabeling every box. Imports append to
an existing database and skip images already in it, so a legacy training set and a new survey's detections can be curated together, and rerunning an import
does not duplicate boxes.

\section{Generate: detection on stills or video}

The detection stage is a thin layer over
Ultralytics YOLO~\cite{ultralytics}, which does all of the model loading,
inference, and multi-object tracking. Supported object-detection models,
including the reader's own fine-tuned weights, can be used; the pipeline expects
box detections and is not a generic wrapper for every task type Ultralytics can
load. Nothing in the pipeline is tied to a particular detector architecture or
class list. Class names are read from the model itself.

Two presets encode the intent of a run. The \code{curate} preset uses a
confidence threshold of 0.005 and a large inference size, and is meant for
building training data: it over-detects on purpose, because a permissive
threshold can be filtered afterward, while a strict one discards detections that
cannot be recovered without a full rerun. The \code{predict} preset uses 0.08 and is meant for routine inference on new imagery. Both are deliberately low starting
points rather than fixed values: confidence, IoU, inference size, and additional hyperparameters are options, and passing one overrides whatever the preset would have supplied. For each detection the pipeline records the box, the class, the confidence, a
blur score computed from the crop, and the crop itself.

A run can also be restricted to a fixed number of images, allowing the pipeline
to be tested on a smaller subset. By default those are the first $N$ files in
sorted order, which is fine for a quick functional test but is not a
representative sample: survey imagery is usually named chronologically, so the
first $N$ files are a contiguous slice of one part of one dive. A seeded random
sample spread across the whole input directory is therefore also available, and
the same directory, count, and seed always select the same images. That option
is what makes the limit useful for right-sizing a real curation effort---a
thousand images spread across a survey---rather than only for testing.

\section{Video: one region of interest per track}\label{sec:video}

Video needs different treatment from stills. An animal in view for three
hundred frames produces three hundred near-identical detections, which
overwhelm clustering, make review tedious, and would inflate counts in any
quantitative analysis. The video command therefore runs an Ultralytics
detector and tracker and keeps one region of interest per track: on the five
one-minute clips described below, 272{,}126 tracked detections became 1{,}163
track representatives. What reaches the reviewer is a list of candidate
observations, not a list of frames.

Because a track's representative cannot be chosen until the track ends, this
runs as two passes. The first tracks the whole video, keeping only a compact
summary of each track, so memory grows with the number of tracks rather than
the length of the video. The second sweeps the file once and writes only the
chosen frames to disk, as ordinary JPEGs. Video-derived rows are then
indistinguishable from image-derived rows downstream, and separate columns keep
the link to the footage, which lets the review GUI open the source video at the
moment of the detection.

The tracker has confidence thresholds of its own, and these, not the detection
threshold, decide which detections form tracks. Ultralytics' shipped
configurations start a track only at confidences of 0.25 to 0.7, far above the
\code{curate} preset's 0.005: on the footage below, fewer than 0.1\% of
detections reached TrackTrack's 0.7, and one clip produced no tracks at all.
The video command therefore defaults to a ByteTrack configuration matched to
the preset (for \code{curate}, a track starts at 0.01 and continues at 0.005),
and warns when a user-supplied tracker's threshold is far above the detection
threshold.

\subsection{Choosing the representative frame}\label{sec:track}

\begin{figure}[H]
\centering
\includegraphics[width=0.98\textwidth]{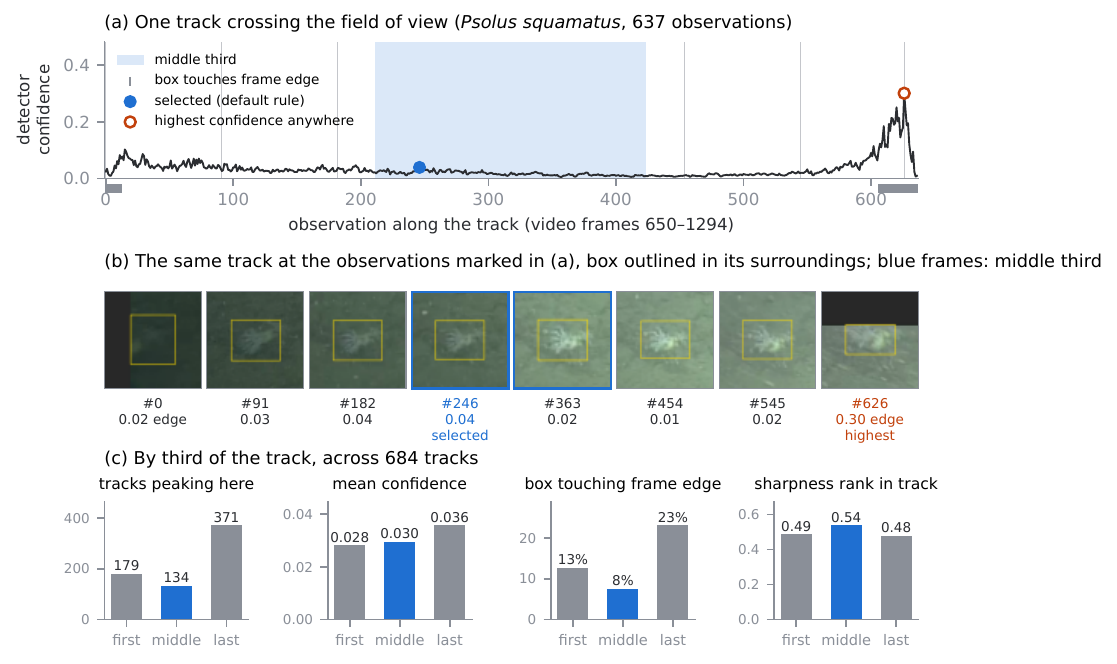}
\caption{Choosing a frame from a track. (a) Detector confidence along one
637-observation track, labeled \emph{Psolus squamatus} by the detector, with
the middle third shaded, boxes touching the frame edge ticked below the axis,
and the frames chosen by the default rule and by highest confidence anywhere
marked. (b) The box in its surroundings at the observations marked in (a);
area beyond the video frame is dark. The animal enters at the left edge, is
whole in the middle third, and is cut off by the top edge in the
highest-confidence frame. (c) Across 684 tracks, by third of the track: where
confidence peaks, mean confidence, the share of boxes touching the frame edge,
and mean within-track sharpness rank. None of these measures is correctness.}
\label{fig:track}
\end{figure}

The software takes the most confident detection from one third of the
track---the first, middle, or last, as the user chooses---or alternatively the
sharpest frame in that third, the most confident frame anywhere, or the center
frame. The middle third is the default, from annotators' experience that an
animal entering or leaving view is often clipped by the frame edge, occluded,
or blurred, and that confidence alone can favor such a frame.
Figure~\ref{fig:track}(a,b) shows a track where this happens: a sea cucumber,
labeled \emph{Psolus squamatus} by the detector, is whole through the middle of
its transit, where the default picks a detection at 0.04, but confidence peaks
at 0.30 as it leaves, clipped by the top edge of the frame.

To look more broadly, we tracked five one-minute clips of seafloor video from a
remotely operated vehicle, using a 415-class YOLO26s detector trained on
MBARI's VARS annotations and the default settings above, and measured every
observation of the 684 tracks at least 30 observations long. A box counted as
touching the frame edge if it ended within 10 pixels of it. Confidence peaked
most often in the last third (371 tracks, against 179 in the first and 134 in
the middle), but the last third also had the most edge-touching boxes---23.1\% of observations, against 12.6\% in the first and 7.5\% in the middle, which also had slightly sharper crops (Figure~\ref{fig:track}(c)). Picking the
most confident frame within each third shows the trade-off: the first, middle,
and last thirds pick an edge-touching box for 20.5\%, 9.2\%, and 37.9\% of
tracks, at a median detector confidence of 0.042, 0.042, and 0.065 (0.086 for
the most confident frame anywhere, which touches the edge for 42.3\%). These patterns
hold at edge margins of 5 and 15 pixels.

These measures are proxies: neither edge contact nor confidence shows whether a
detection is correct or whether the frame is the one an expert would choose.
They support the reasoning behind the middle-third default without showing that
it yields better training data, so the choice of third is left to the user. A
seeded sample of 24 tracks, inspected from contact sheets by an AI model (Claude
Opus~5.5) rather than by a human expert, each followed a single object
throughout; fragmentation of one animal into several tracks was not assessed.

\section{Enrich: embeddings and clusters}

Clustering is optional, but it has the potential to greatly speed up the review
process. Using DINOv3 (ViT-L/16, \code{vit\_large\_patch16\_dinov3.lvd1689m} via
\code{timm}), an embedding vector is generated for each ROI. These embeddings are clustered with EVoC, and each cluster is then labeled with the dominant class
among its members, which propagates one decision across everything in the group.
A reviewer looking at a page of visually similar sponges can accept them as a
batch instead of singly. Because a single embedding model defines the geometry of
that space, mixing vectors from two models in one database would make the
distances meaningless; changing the model therefore requires recomputing every
embedding rather than only the missing ones.

The same embeddings drive a similarity search in the GUI. Right-clicking any
ROI re-ranks the entire database by cosine similarity to it,
which turns ``show me the rest of these'' into a single action. Embeddings are
mean-centered on the ranking pool before normalization.

Naming the clusters turns out to need care. The obvious approach, and the one
we started with, is to label a cluster after the dominant original class among
its members. That assumes the detector has enough classes for the label to
distinguish one cluster from another, which is not always true. A single-class
detector---MBARI's Megalodon, for instance, which reports only \code{object}---makes every cluster's dominant label the same string, so writing it back
collapses the clustering that was just computed into one undifferentiated
label. The structure survives only in the raw cluster number, and the review
GUI, which filters and sorts on the curated label, can no longer separate the
groups at all.

Clusters are therefore named with an index appended whenever one label wins
more than one cluster: a single-class run yields \code{object\_1},
\code{object\_2}, and so on. On a 147-region run from a single-class model,
six clusters that previously all became \code{object} now come back separately
labeled. The same rule helps multi-class runs, where several clusters of the
same taxon would otherwise be flattened together; on the same regions with the
detector's real labels, two singleton clusters keep their bare names while four
distinct sponge clusters become \code{Hexactinellida\_1} through
\code{\_4}. Merging them afterward is one bulk relabeling command, whereas
recovering a distinction that was thrown away is not.

These suffixed names encode review groups, not biological classes. They are
provisional labels used to keep clusters separate in the current flat schema.
Before producing a training export, a reviewer consolidates them into the
appropriate taxonomic label or replaces them with a confirmed label; the suffix
itself should not become a model class.

A related choice is which label the naming vote reads. Voting on the raw
detector class is right for a first pass over fresh detections, but wrong for a
database that has already been reviewed: there the raw class is often the same
uninformative string on every row, while the names a human typed are the ones
worth carrying onto the new clusters. The naming vote can therefore read either
the raw class or the curated label, falling back to the raw class for rows not
yet reviewed.

Clustering is a destructive write and cannot be undone. A clustering run overwrites the curated label of every embedded row, including rows a human has already verified, and it does not clear
the verified flag---so afterward those rows still present as human-reviewed while carrying a machine-generated name. Experiments on a curated database should be run against a copy. For the same reason, naming from the curated label is a first-pass operation: run twice against the same file, the second pass votes on the names the first pass generated.

When the first clustering pass lumps things too coarsely, a refine step
re-clusters a single label into finer groups, which is the usual remedy for a
large ambiguous cluster.

\section{Review: the human in the loop}

\begin{figure}[H]
\centering
\includegraphics[width=\textwidth]{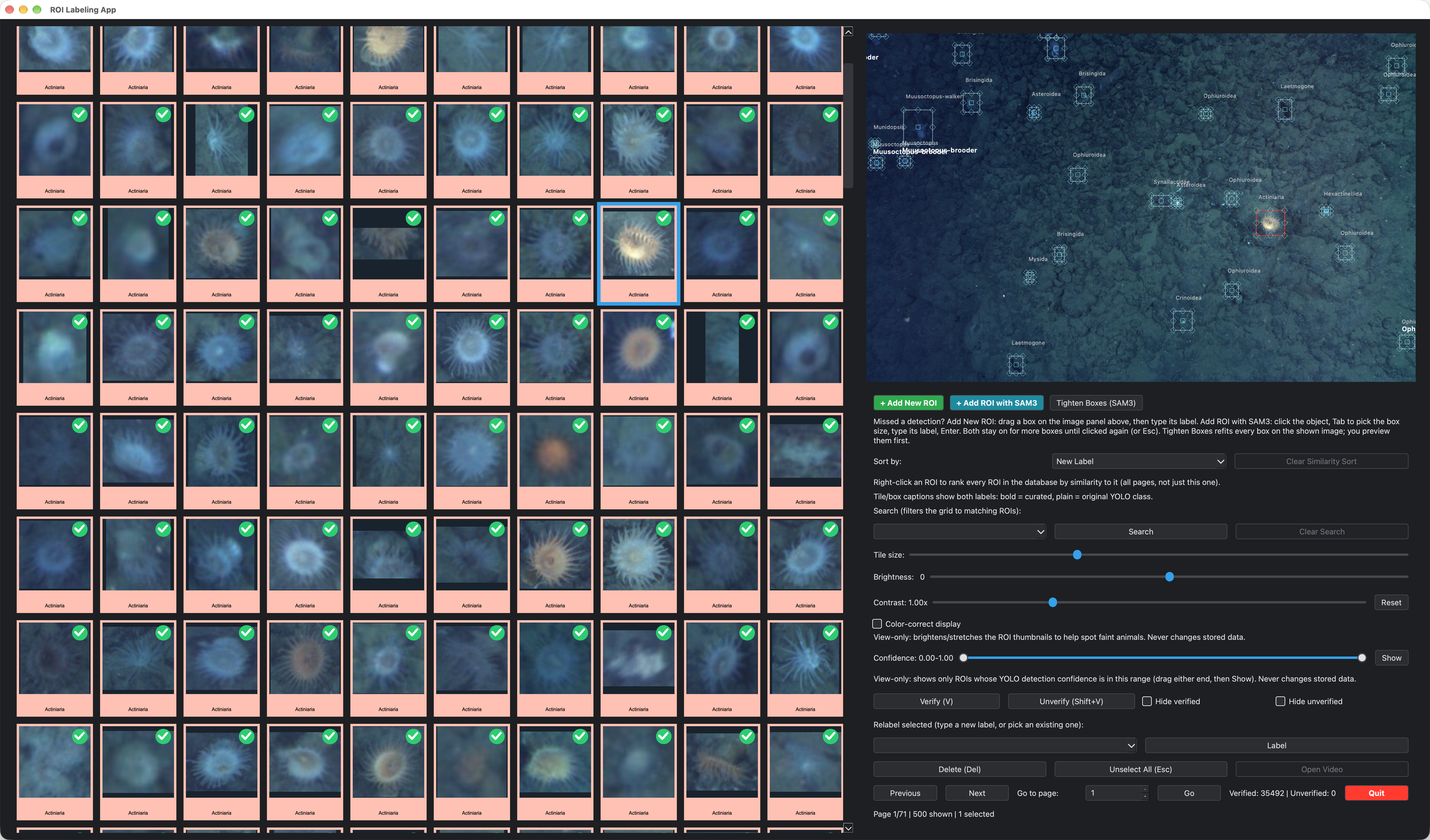}
\caption{The review GUI. Left: the mosaic of regions of interest, tinted by
label, with a badge on those already checked and the selected region outlined.
Right: the full source frame with every detection on it drawn as an editable
box, the selected one in red, above the controls. Brightness, contrast, and
color correction for the blue-green water cast change only how the imagery is
displayed, never the stored data, which matters for faint animals against
sediment.}
\label{fig:gui}
\end{figure}

The review GUI (Figure~\ref{fig:gui}) presents regions of interest as a grid of thumbnails, tinted by label, with a checkmark identifying those that have been
validated. Original and curated labels are both shown on each tile, and the grid
can be sorted by either, or by image name, sharpness, or aspect ratio;
right-clicking a region re-ranks by similarity to it instead. The grid can also
be filtered by label, by a band of detector confidence (both ends adjustable,
so that, for example, only the 0.2--0.3 band can be reviewed), and by review
state, hiding either the verified or the unverified regions. A label filter
matches each region's effective label, the same \code{COALESCE(new\_label,
label)} rule the exports use, so a region renamed away from a class no longer
appears under it. Filters are applied to the
whole database rather than to the page on screen, so the reviewer is always
paging through every region that matches, not through whatever happened to be
loaded. An annotator working through the review GUI can:

\begin{itemize}\itemsep1.5pt
\item \textbf{Validate} a detection, marking it reviewed. Applying a label counts
      as review, since labeling something is itself an act of checking it.
\item \textbf{Relabel} one region or a selection of hundreds, either by typing a
      new concept or picking one already in use in this database.
\item \textbf{Adjust} a box by dragging its handles on the full-resolution frame.
      The stored crop and blur score are regenerated to match, immediately,
      and the embedding is recomputed in the background, so that similarity
      search and clustering place the region by what it now contains rather
      than by its old crop.
\item \textbf{Delete} false positives, individually or in bulk.
\item \textbf{Add} a localization the model missed entirely, by drawing a box on
      the frame and naming it. Its embedding is computed in the background using
      the same model and preprocessing as the batch path, so a hand-drawn region
      is directly comparable to every other one and participates in similarity
      search straight away. A box can be drawn inside or on top of an existing
      one, for a part within a whole or an animal in front of a rock.
\item \textbf{Add with SAM3}, optionally: a single click on an object prompts
      the SAM3 segmentation model~\cite{sam3} for its box. SAM3 usually offers
      nested candidates (a part, the object, the object with its
      surroundings), which the reviewer steps through before naming one.
\item \textbf{Tighten boxes with SAM3}, optionally: every box on the shown frame
      is refitted from SAM3's mask and previewed, and nothing is written until
      the reviewer confirms (Section~\ref{sec:sam3}).
\end{itemize}

The right-hand panel draws \emph{every} detection on the source frame, not only the
ones on the current page, because detections from the same frame are often scattered across different pages by sorting. Without this, a reviewer editing a box has no idea what else was found on that image.

Deep-sea imagery is frequently dark, low in contrast, and tinted blue-green by
the water, which hides faint animals and drains the color that often
separates one taxon from another. The GUI provides three display controls for
this, alongside a tile-size slider: brightness and contrast for the thumbnails,
and an optional color correction applied to both the thumbnails and the full
frame. The color correction is measured once per source image, from a reduced
copy of the whole frame, so a crop keeps its color relative to its
surroundings: each channel is corrected half by a 0.5--99.5 percentile stretch,
which restores contrast and the weak red channel, and half by a gray-world
balance, which removes the cast; contrast-limited adaptive histogram
equalization on lightness and a modest saturation boost follow. Per-image
measurements are cached beside the database. All three controls change only
what is displayed: labels, boxes, embeddings, and the crops regenerated after
a box edit are always computed from the original pixels.

Similarity search reports its own coverage; the status bar states how much of the matching set was actually ranked. Display filters---review state, label, and confidence band---are applied in the query that builds each page rather than
to the page after it is built, so that paging through a filtered database visits
only rows that pass the filter.

A separate bulk-relabeling command handles taxonomy revisions across the
entire database, applying all renames in a single pass so that a file that swaps
two labels does what it says.

\subsection{SAM3-assisted boxes}\label{sec:sam3}

SAM3~\cite{sam3}, run through Ultralytics, is optional: without it the two SAM3
controls are disabled, with the reason given, and the GUI behaves exactly as
before. With it, a click becomes a box, and existing boxes can be tightened.
Tightening matters most for imported and hand-drawn boxes, which usually carry
a margin, and for detector boxes that include surrounding sediment. Each box
is prompted with the box itself and a positive click at its center. The new
box is measured from SAM3's most confident mask, counting only mask pixels
within the original box (plus 5\% on each side) and ignoring isolated specks;
that box becomes the next prompt, for up to four prompts. Prompting with the
box alone often let the mask spill onto the seafloor, so that boxes stayed loose,
grew, or drifted sideways.

On 60 imported marine-litter boxes, tightening left a median 0.45 of the original box area, and in 2 of the 60, SAM3 appeared to segment something other
than the boxed object; those boxes are left unchanged rather than offered.
Taking SAM3's smallest mask instead of its most confident one gave tighter boxes
still, but cut off parts of objects, such as a tire's rim or a bottle's cap.
These figures describe how much tightening changes a box, not whether the
result is correct, and a tighter box is not always a better one: on hand-drawn
boxes SAM3 occasionally removed a thin or faint extremity, such as the far end
of a cable or a fish's tail. Every proposed change is therefore previewed on
the frame, and nothing is written until the reviewer accepts it. Tightening
takes about 0.6\,s per box on an Apple M3 Ultra, image features included.

\section{Export}

The export stage writes Pascal VOC XML, YOLO-format label files with the
accompanying names file, \code{*.id} sidecar files for MBARI's own tooling, and a
paginated HTML gallery for quick visual checks. All of them select the same
population by the same rule by default---verified regions, by effective
label---and all of them locate imagery through the path
recorded at ingest, so none takes a separate image-directory argument. Every
export except VOC can be asked to include unreviewed detections as well, still
excluding \code{noise}, for example to bootstrap a first dataset from model
output. The VOC
and YOLO exports also emit a manifest of the source images and a small
standalone script that can be used to copy them to the desired location; the
script depends only on the Python standard library, so it runs on a machine
where the pipeline itself is not installed. For a YOLO export the script
requires its destination to be named, because the images are only usable in
the \code{images/} directory beside the label files that the split lists
refer to.

\subsection{Output filenames}

Every export that writes one file per source image names it after that image's
own filename. There is one exception, and it exists because flattening a nested
mission directory tree into a single output directory is precisely the situation
in which two dives' identically named images become one output file, the second
silently overwriting the first. Naming is therefore decided across an export as
a whole: filenames that are unique are used as they are, and only those that
actually collide fall back to a parent-directory prefix, with the collision
reported. One unlucky pair in a large survey does not rename everything else,
and an export whose names cannot be made unique fails rather than overwriting.

\subsection{YOLO datasets}

The YOLO export writes a dataset that is ready for training. It contains a
label file for each image, the class names in index order, lists of training,
validation, and test images, and the configuration file the Ultralytics trainer
reads.

The split is made over images, not regions of interest. The list of exported
images is shuffled and divided by the chosen ratios (85/10/5 by default), and
each image's regions go with it, so no image or region appears in more than one
split. Splitting regions instead would let crops of the same frame, often of the
same animal, fall on both sides of the train/validation boundary and inflate
validation scores. The shuffle is seeded, so exporting the same images again
reproduces the same split; exporting a different set of images, for example
after more have been reviewed, produces a new one. Images that must stay out of
training across exports, such as a fixed benchmark set, can be pinned into the
test split.

The class names in the configuration come from the same list that assigned the
class indices in the label files, so the two cannot disagree. Paths in the
configuration are relative to the file itself, so the dataset directory can be
moved or copied to a training machine without editing, and an empty split is
left out of the configuration rather than pointing at an empty file.

Two options help when adding to an existing dataset. The first keeps that
dataset's class order: given its names file or configuration, every class keeps
its index, and any new labels are added at the end with a warning. The second is
a confidence threshold for unreviewed detections, applied to whole images: an
image is exported only if all of its boxes meet the threshold. Dropping just the
low-confidence box would leave that object unlabeled in the image, and YOLO
would learn it as background. Verified boxes always pass, since a person has
confirmed them.

\subsection{Identification sidecars}

The \code{*.id} sidecar is the interchange format for MBARI's own tooling, and
its layout is designed for a downstream process rather than for a reader. Each
file carries a commented header---the tool version, the operator, the time, the
model, the full path of the source image, and the image's pixel dimensions---followed by one comma-separated row per identification:

\begin{quote}\footnotesize\ttfamily
index,\,label,\,confidence,\,center\_x,\,center\_y,\,lon,\,lat,\,depth,\\
tl\_x,\,tl\_y,\,tr\_x,\,tr\_y,\,br\_x,\,br\_y,\,bl\_x,\,bl\_y
\end{quote}

Each observation carries both a single position---the box's center pixel---and
the box itself as four corners, since consumers want one or the other and
deriving one from the other independently is how two consumers come to disagree.
The center is computed as the integer midpoint of the corners as written, not of
the underlying floating-point values, so a reader that derives it obtains
exactly the value in the file. The geographic fields are placeholders: a
separate navigation process populates latitude, longitude, and depth for each
center point by rewriting these files, and the header names those three columns
as the only ones it should alter.

By default the sidecars are written beside their source images, following the
mission's own directory structure; they can also be exported into a user-selected
directory.

\subsection{Summaries and ad hoc questions}

Two commands support analysis rather than training. A statistics subcommand of
the export group reports label counts and per-image detection counts, and writes a label-by-image matrix suitable for ecological work such as per-image richness or frequency of
occurrence. Because it applies the same selection rule as the annotation
exports, its counts are a preview of what a training set will contain rather
than a different measurement of the same database; both it and the HTML gallery
can be asked to include unreviewed detections instead, which is what makes them
useful against a database fresh out of inference. An ad hoc query command can be used to run arbitrary SQL against the database, which is often the fastest way to answer a question the tool does not anticipate.

\section{Limitations}

This paper reports an internal data-reduction example and exploratory
observations, not a controlled evaluation of detection accuracy, annotation
throughput, or frame-selection quality. The track-selection analysis measures
proxies---confidence, edge contact, and sharpness---on five clips from one kind
of footage with one detector, and the rule's benefit for review or training has
not been quantified. The longest video
processed end to end is short; the first pass retains compact per-track metadata
rather than frames, but its scaling is unverified at full dive length.
Clustering quality depends on EVoC parameters that in practice need tuning per
deployment. Track identifiers are also imperfect proxies for individual
animals: missed detections and occlusions can fragment one animal into multiple
tracks, while identity switches can join observations of different animals.
Track representatives must therefore be treated as candidates for human review,
not as guaranteed unique individuals. SAM3 tightening was assessed only by how
much it shrinks boxes and by inspection, on one marine-litter dataset, and not
against independent ground truth; it remains a proposal the reviewer accepts or
rejects. Because DuckDB takes an exclusive lock on the database file, two people cannot curate the same database at once, and combining separately curated
databases is a manual merge, although importing one database's exported
annotations into another is now possible.

\section{Availability and credits}

\code{mbariml} is written in Python and available under the MIT license at
\url{https://github.com/mbari-org/mbari-ml}. The repository carries the
reference documentation for every command, a cheat sheet covering the
walkthrough in this paper, a schema reference for the database described in Section~2, and a changelog recording the measurements quoted here alongside the
changes that produced them. Every command's own \code{-{}-help} output is
generated from the code and is authoritative where it and the prose differ.

Detection and tracking are provided entirely by \textbf{Ultralytics
YOLO}~\cite{ultralytics}, which the pipeline wraps rather than reimplements;
model loading, inference, and the multi-object trackers are all theirs, as is
the interface through which the optional SAM3 model~\cite{sam3} is run. The
review GUI's mosaic rendering, threading, and box-overlay machinery are adapted
from MBARI's \emph{vars-gridview}~\cite{varsgridview} under the MIT license.
Embeddings use DINOv3~\cite{dinov3} via \code{timm}, clustering uses
EVoC~\cite{evoc}, and storage is DuckDB~\cite{duckdb}. The database layer, the
dataset import, the video ingest and track-reduction stages, the export
formats, and the additions to the review GUI described here are original to
this work.

\section*{Acknowledgments}

\noindent\textbf{AID Statement:} \textbf{Artificial Intelligence Tool:}
ChatGPT-5.6 Sol Medium, Claude Sonnet 5 High, Claude Opus 5, Claude Opus 5.5; \textbf{Information
Collection:} Claude Sonnet 5 High and Claude Opus 5 were used to review the open-source
codebase being described; \textbf{Data Collection Method:} Claude Sonnet
5 High and Claude Opus 5 were used to review the open-source codebase being
described; \textbf{Data Analysis:} Claude Sonnet 5 High and Claude Opus 5 were
used to test the code described in the manuscript on a test dataset;
\textbf{Interpretation:} Claude Sonnet 5 High and Claude Opus 5 were used to
interpret the results of the code tests; \textbf{Writing -- Review \& Editing:}
Claude Sonnet 5 High and Claude Opus 5 were used to assist in writing the final
paper. Claude Sonnet 5 High, Claude Opus 5, and ChatGPT-5.6 Sol Medium were used
to help edit and make final revisions. Claude Opus 5.5 was used to rerun the
track-selection analysis and produce Figure~\ref{fig:track}, to inspect a
sample of the tracks from contact sheets, and to revise the text for the
software's later additions.

\end{document}